\documentclass[]{llncs}

\usepackage{graphicx}
\usepackage{diagbox}
\usepackage{algorithm}
\usepackage{algorithmic}
\usepackage{boldline}
\usepackage{amsfonts}
\usepackage{amssymb}
\usepackage{mathtools}
\usepackage{multicol}
\usepackage{url}
\usepackage{booktabs}

\usepackage{pifont}

\usepackage{multirow, multicol}

\begin{document}
\title{Semantic Pivoting Model for Effective Event Detection\thanks{Supported by the Agency for Science, Technology and Research, Singapore.}}
%
%
\author{Hao Anran\inst{1,2} \and
Hui Siu Cheung\inst{1} \and
Su Jian\inst{2}}
\authorrunning{A. Hao et al.}
%
\institute{Nanyang Technological University, Singapore\\ \email{\{S190003, asschui\}@ntu.edu.sg}\\\and
Institute for Infocomm Research, Agency for Science, Technology and Research, Singapore\\
\email{sujian@i2r.a-star.edu.sg}}
\maketitle              
\begin{abstract}
Event Detection, which aims to identify and classify mentions of event instances from unstructured articles, is an important task in Natural Language Processing (NLP). Existing techniques for event detection only use homogeneous one-hot vectors to represent the event type classes, ignoring the fact that the semantic meaning of the types is important to the task. Such an approach is inefficient and prone to overfitting. In this paper, we propose \underline{S}emantic \underline{P}ivoting Model for \underline{E}ffective \underline{E}vent \underline{D}etection (SPEED), which explicitly incorporates prior information during training and captures more semantically meaningful correlation between input and events. Experimental results show that our proposed model achieves the state-of-the-art performance and outperforms the baselines in multiple settings without using any external resources.


\keywords{Event detection \and Information extraction \and Natural Language Processing \and Deep learning}
\end{abstract}

\section{Introduction}

Event Detection (ED), which is a primary task in Information Extraction, aims to detect event mentions of interests from a text. ED has wide applications in various domains such as news, business and healthcare. It also provides important information for other NLP tasks including Knowledge Base Population and Question Answering. The state-of-the-art ED models are predominantly deep learning methods, which represent words using high dimensional vectors and automatically learn latent features based on training data \cite{chen-etal-2015-event,25years-of-ie-grishman2019}. However, limited size and data imbalance of ED benchmarks pose challenges in performance and robustness of current deep neural models \cite{chen-etal-2017-automatically}. For instance, over 60\% of the types in the ACE 2005 benchmark dataset have less than 100 data instances each.

Recent works on ED can be categorized into three major approaches: (i) proposing architectures with more sophisticated inductive bias \cite{chen-etal-2015-event,yan-etal-2019-event}; (ii) leveraging on linguistic tools and knowledge bases \cite{lu-nguyen-2018-similar}; (iii) using external or automatically augmented training data \cite{wang-etal-2019-adversarial-training}. These approaches can be seen as indirectly alleviating the lack of type semantic prior in the model, but they ignore the important fact that the types are semantically meaningful. The models only treat each event type class homogeneously as one-hot vectors and are therefore agnostic to the semantic difference or association of the types.


In this paper, we propose to directly incorporate the type semantic information by utilizing the class label words of the event types (e.g.,   ``attack" and ``injure") to guide ED. To this end, we leverage the state-of-the-art Language Model (LM) structure, Transformer \cite{transformer-vaswani2017attention}, and propose a Semantic Pivoting Model for Effective Event Detection (SPEED), which uses the event type label words as auxiliary context to enhance trigger classification through a two-stage network. We highlight the fact that the label words are natural language representation of the meanings of the target types, which allows us to: (1) use them as initial  semantic pivots for ED, and (2) encode them in the same manner as the input sentence words and enhance the representations of both via the attention mechanism.




To the best of our knowledge, this is the first work to exploit the event class label set and incorporate the type semantic prior information for the task. We evaluate our SPEED model on ACE 2005 benchmark and achieve the state-of-the-art performance. The rest of the paper is organized as follows: Section~\ref{sec:related-work} reviews the related work and Section~\ref{sec:preliminaries} specifies the task formulation. In Section~\ref{sec:proposed-method}, we introduce our proposed SPEED model. Section~\ref{sec:experiments} discusses the experimental details including the dataset, compared baseline models, hyperparameter settings, performance results and analysis. Section~\ref{sec:conclusion} gives the conclusion of this paper.

\section{Related Work}
\label{sec:related-work} 

Deep learning models \cite{chen-etal-2015-event,nguyen-grishman-2015-event} which are based on distributed vector representations of the text and neural networks have been widely used for modern ED. Such approaches automatically extract latent features from data and are thus more flexible and accurate than early feature-based approaches (e.g., \cite{liao-grishman-2010-using}). Over the years, other than convolutional neural network (CNN) \cite{chen-etal-2015-event} and recurrent neural network (RNN) \cite{nguyen-grishman-2015-event}, more sophisticated architectures or mechanisms including attention \cite{liu-etal-2017-exploiting}, Transformer and graph neural network \cite{liu-etal-2018-jointly,wang-etal-2019-adversarial-training,yan-etal-2019-event} are introduced to improve the performance. However, the data scarcity and imbalance problems remain as the bottleneck for substantial improvement. To alleviate the data scarcity problem, many works leverage on external linguistic resources such as Freebase and Wikipedia \cite{chen-etal-2017-automatically,wang-etal-2019-adversarial-training} to generate auto-augmented data via distant supervision. Utilization of pre-trained language models, joint extraction of triggers and arguments, and the incorporation of document-level or cross-lingual information \cite{du-cardie-2020-event,subburathinam-etal-2019-cross,wadden-etal-2019-entity} are also found to be able to enhance ED.


Class label representation has been used for image classification, but rarely explored for natural language processing (NLP) tasks. The recent works \cite{wang-etal-2018-joint-embedding,zhang-etal-2018-multi} encoded label information as system input for Text Classification. Recently, Nguyen et al. \cite{nguyen-etal-2019-employing} demonstrated the effectiveness of explicitly encoding relation and connective labels for Discourse Relation Recognition. However, these methods learn separate encoders for the labels and the input sentence words. This is redundant because the words used in both the labels and the sentences are from the English vocabulary and they can share the same embedding. Furthermore, these methods do not effectively model the rich interactions between sentence words and labels as well as between two event labels. In our work, the label input shares the distributed representation with the input text, while the deep attention-based structure captures higher-order interactions among word and label tokens. By harnessing the power of pre-trained language models, we avoid the hassle of data augmentation methods such as distant supervision \cite{muis-etal-2018-low}, in which much noise is introduced.

\section{Event Detection}
\label{sec:preliminaries}

ED is formulated as identifying \textit{event triggers} which are the words that best indicate mentions of events, and classifying these triggers into a pre-defined set of \textit{event types}. For example, in the sentence S1, the underlined words are the triggers of an \texttt{Attack} event and an \texttt{Injure} event respectively:

\begin{quote}
    \textbf{S1:} A bomb \underline{\textbf{went off}} near the city hall on Friday,  \underline{\textbf{injuring}} 6.
\end{quote}

\noindent We formulate the task as a word-level sequence tagging problem, with the input being the document sentences and the output being the predicted trigger type labels of each word spans. We follow the criteria used in previous ED works \cite{chen-etal-2015-event,li-etal-2013-joint,nguyen-etal-2016-joint} and consider an event trigger as correct if and only if both the boundary and classified type of a trigger match the corresponding ground truth.

\section{Proposed Model}
\label{sec:proposed-method}

\begin{figure*}[t]
\centering
\includegraphics[width=160pt]{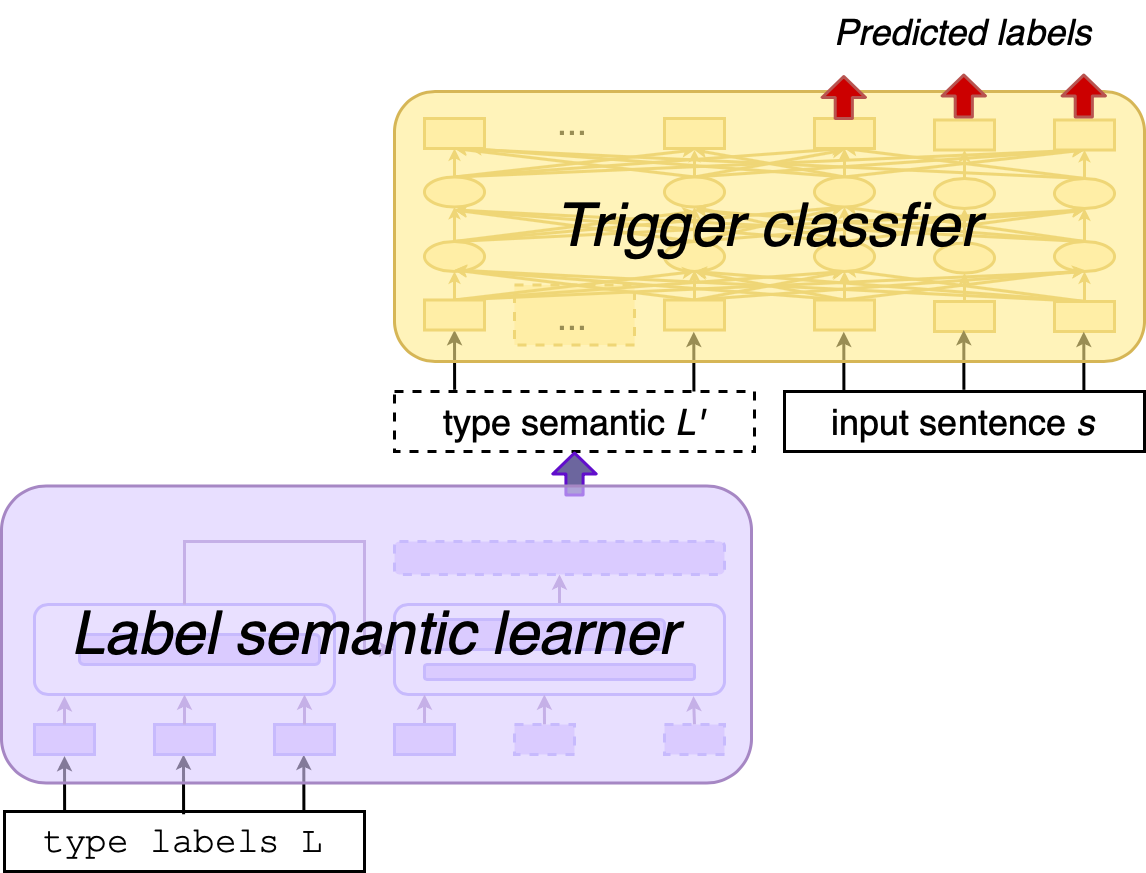}
\caption{Architecture of the proposed \textsc{Speed} model.}
\label{fig:overall}
\end{figure*}

Figure~\ref{fig:overall} shows the overall architecture of our proposed SPEED model, a two-stage Transformer-based model consisting of a Label Semantic Learner and a Trigger Classifier.


\subsection{Label Semantic Learner}
\label{sec:component1}

\begin{figure*}[t]
\centering
\includegraphics[width=\linewidth]{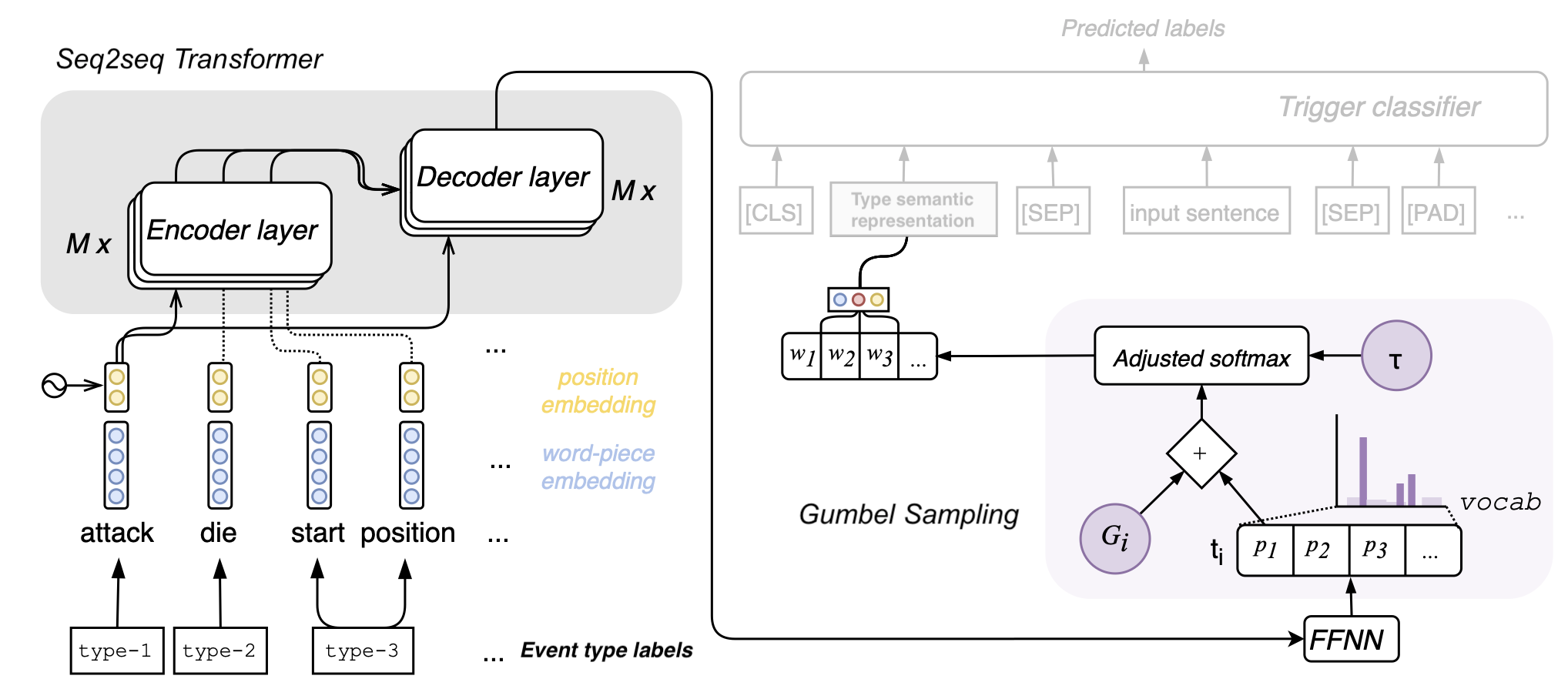}
\caption{Architecture of Label Semantic Learner.} \label{fig:ss}
\end{figure*}

Figure~\ref{fig:ss} shows the proposed Label Semantic Learner, which employs Sequence-to-sequence (Seq2seq) Transformer and Gumbel Sampling.

\paragraph{Sequence-to-sequence Transformer.} To learn a semantic representation of event types based on the label words, we first concatenate words in the original event type labels, forming a sequence $L = \langle l_1, ..., l_n\rangle$. Then, we randomly shuffle the labels to reduce the influence of positional embedding. The Label Semantic Learner takes the label words as input and passes the representation through the Transformer architecture \cite{transformer-vaswani2017attention}, which consists of $M$ encoder layers followed by $M$ decoder layers. The attention mechanism within each layer allows the type label words to interact with each other based on lexical semantic similarity and difference. Note that the encoder and decoder attention masks are set to allow each token $l_i \in L$ to interact with all other tokens either before or after it. The final decoder layer is connected to a feed-forward neural layer (\textit{FFNN}), which predicts a new sequence $L^\prime = \langle w_1, ..., w_n\rangle$ that encodes type semantic information. To be consistent with the original label sequence, we restrict the number of tokens in the output sequence to form $L^\prime$ to be the same as that of the input $L$.
 
\paragraph{Gumbel Sampling.} The next step of the Label Semantic Learner infers the best-suited label words from the vocabulary based on the distribution. This involves ``discrete" steps of taking the most possible next tokens, causing the backpropagation problem. Also, there are conceivably multiple ways to describe the meanings of the event types. Instead of deterministically choosing the word of highest probability, the model may benefit from a ``softer", probabilistic approach that also allows other words to be chosen (those with lower probabilities). Thus, we employ the Gumbel-Softmax method \cite{gumbel-softmax-jang2016categorical}, which closely approximates the \textit{argmax} operation via Gumbel Sampling, for the Label Semantic Learner.

More specifically, we replace the usual non-differentiable token prediction operation that selects one word $w_i$ from the vocabulary with the highest probability:
\begin{equation}
\label{eq:1}
w_i=argmax{(softmax{(p_i))}}
\end{equation} with:
\begin{equation}
\label{eq:2}
w_i=argmax{(softmax{(\frac{p_i+G_i}{\tau}))}}
\end{equation} where $p$ are the computed probability logit values of $w$, $\tau$ is the temperature parameter controlling the degree of approximation and $G_i$ is a random noise value sampled from the \textit{Gumbel-Softmax} distribution $G$:
\begin{equation}
G = -\log(-\log(U_i)), U_i \sim Uniform{(0,1)}
\end{equation}
This basically reparameterizes $p_i$ and replaces the sample $w_i$ in Equation (1) drawn from the one-hot-encoded, categorical distribution on the vocabulary with an approximation drawn from a continuous distribution ($G$) as in Equation (2), thereby allowing the backpropagation to compute the respective gradient.

\subsection{Trigger Classifier}
\label{sec:component2}

\begin{figure*}[t]
\centering
\includegraphics[width=.8\linewidth]{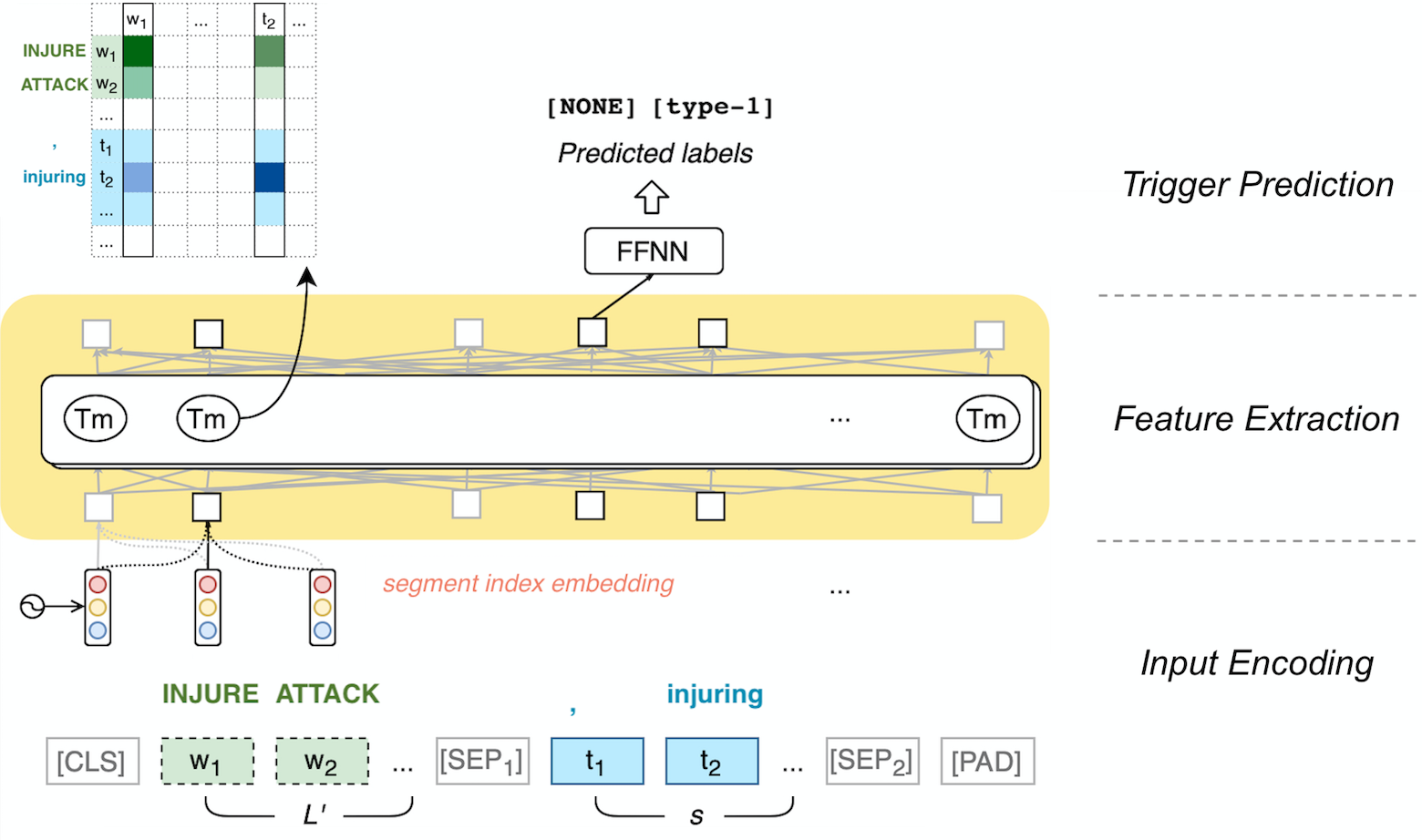}
\caption{Architecture of Trigger Classifier.} \label{fig:ed}
\end{figure*}

Figure~\ref{fig:ed} shows the architecture of Trigger Classifier, which consists of Input Encoding, Feature Extraction and Trigger Prediction.

\paragraph{Input Encoding.} We leverage the state-of-the-art Pre-trained Language Model BERT \cite{transformer-vaswani2017attention} for encoding. As shown in Figure~\ref{fig:ed}, we construct the input by concatenating each sentence with the type semantic sequence. After adding special tokens in BERT including [CLS] and [SEP], for each input sentence $s$ (of length $N_S$) with all the label words $L^\prime$ (of length $N_{L^\prime}$), the input sequence is as follows:
\begin{equation}
  X_{s} = \langle [\textsc{CLS}], L^\prime, [\textsc{SEP}_1], s, [\textsc{SEP}_2] \rangle
\end{equation} of length $N_{valid}$ = $N_s + N_L^\prime + 3$. 

Following BERT, three types of embeddings are used, namely word-piece embedding $E_w$, position embedding $E_p$ and segment index embedding $E_s$. For each word-piece token $x_i \in X_s$, it is embedded as:
\begin{equation}
    e_i = E_w(x_i) \oplus E_p(x_i) \oplus E_s(x_i)
\end{equation}
\[    E_s(x_i) = 
   \begin{cases} 
      f_s(1) & x_i \in L^\prime \textup{ or } x_i=\textsc{SEP}_2 \\
      f_s(0) & \textup{otherwise}
   \end{cases}
\]

\noindent where $f_s$ is the pre-trained segment index embedding from BERT.

\paragraph{Feature Extraction.} At each Transformer layer, contextualized representation for tokens is obtained via aggregation of multi-head attention (denoted by `Tm' in Figure~\ref{fig:ed}):
\begin{equation}
    \textsc{MultiHead}(Q, K, V ) = \langle \text{head}_1, ..., \text{head}_H \rangle W^O
\end{equation}
\[    \text{head}_i = f_\textsc{Attn}(Q W_i^Q , K W_i^K , V W_i^V ), i \in \text{[}1,H\text{]}
\]
where $H$ denotes the number of heads and $Q, K, V$ are query, key, value matrix respectively. In our proposed model, candidate words from the input sentence and words from the type semantic sequence are jointly modeled in the same vector space. The \textsc{Attn} of each head can be decomposed as follows:
\begin{displaymath}
    \textsc{Attn}(Q, K, V) \Rightarrow
       \begin{cases} 
      \textsc{Attn}(Q_s, K_s, V_s) \\
      \textsc{Attn}(Q_{L^\prime}, K_{L^\prime}, V_{L^\prime})  \\
      \textsc{Attn}(Q_{L^\prime}, K_s, V_s)  \\
      \textsc{Attn}(Q_s, K_{L^\prime}, V_{L^\prime})  \\
   \end{cases}
\end{displaymath}
These correspond to four types of token interactions. The first one is the interaction between each pair of input sentence words, capturing solely sentence-level contextual information. The second is the interaction between each pair of type label words, which models the correlation between event type labels. The remaining two allow $s$ to be understood in regards to $L^\prime$ and vice versa.

For example in Figure~\ref{fig:ed}, each word $t_i$ in the example sentence \textbf{S1} ($s$) is a candidate trigger word, and the two event types, among others, are part of the semantic type representation $L^\prime$. Suppose the true label of the token $t_2$ (\textit{``injuring"}) in the input sentence $s$ is semantically represented by $w_1$ (\texttt{Injure}). In our model, the representation of type semantic word $w_1$ at each layer is enriched by that of a similar or contrary type semantic word, such as $w_2$ (\texttt{Attack}). Moreover, it is substantiated by the input sentence words $t_i \in s$, especially the trigger word candidate token $t_2$ as it occurs to be an instance of $w_1$. For the input sentence token $t_2$, its representation is contextualized by other tokens in the same sentence ($t_i \in s$), whereas its attentions with $w_1$ and other label semantic tokens ($w_i \in L^\prime$) provide semantic clues for its ED classification.

\paragraph{Trigger Prediction.} Finally, a feed-forward neural layer predicts an event type $\hat{y}_i$ for each input sentence token $t_i$: $\hat{y}_i = \textsc{FFNN}(t_i)$.

\section{Experiments}
\label{sec:experiments}

This section discusses the dataset, evaluation metrics, baseline models for comparison and experimental results.

\subsection{Dataset and Evaluation Metrics}

We conduct the experiments based on ACE 2005, a benchmark dataset for Event Detection and the most widely-used ED benchmark to date. The documents are gathered from six types of media sources: newswire, broadcast news, broadcast conversation, weblog, online forum and conversational telephone speech. The annotation includes 33 fine-grained event types. We evaluate our models on its English subset. We use the same split as in the previous ED work \cite{chen-etal-2015-event,wang-etal-2019-adversarial-training,yan-etal-2019-event}. The details of the dataset and split are summarized in Table~\ref{table:stats}. \textit{Eventful} sentences refer to those with at least one event mention. For the evaluation, we report the precision (\textbf{P}), recall (\textbf{R}) and micro-average \textbf{F1} scores.

\begin{table}[t]
\centering
\caption{\label{table:stats}Data split and statistics.}
\resizebox{.6\linewidth}{!}{%
\begin{tabular}{lcccc}
\hline
      & \textbf{\# Docs} & \textbf{\# Sents} & \textbf{\# Eventful} & \textbf{\# Triggers} \\
\hline
\textbf{Train} & 529     & 14347    & 3352              & 4420        \\
\textbf{Dev}   & 30      & 634      & 293               & 505         \\
\textbf{Test}  & 40      & 840      & 347               & 424    \\\hline    
\end{tabular}
}
\end{table}

\subsection{Baselines}

We compare the performance of our model with three kinds of baselines: (1) models that do not use linguistic tools or extra training data; (2) models that use linguistic tools to obtain auxiliary features such as POS tag, dependency trees, disambiguated word sense; and (3) models that are trained with extra data. The baseline models are discussed as follows:

\begin{itemize}
    \item \textbf{DMCNN} \cite{chen-etal-2015-event} is a  CNN-based model that uses dynamic multi-pooling.
    \item \textbf{DMBERT, DMBERT+Boot} \cite{wang-etal-2019-adversarial-training} have a pipelined BERT-based architecture for ED. DMBERT+Boot is DMBERT trained on an augmented dataset \textit{Boot} from external corpus through adversarial training.
    \item \textbf{BERT\_QA} \cite{du-cardie-2020-event} performs ED in a QA-like fashion by constructing generic questions to query BERT.
    \item \textbf{JRNN} \cite{nguyen-etal-2016-joint} is a RNN-based model for joint ED and argument extraction.
    \item \textbf{JMEE} \cite{liu-etal-2018-jointly} jointly extracts event triggers and event arguments with a Graph Convolutional Network (GCN) based on parsed dependency arcs.
    \item \textbf{MOGANED} \cite{yan-etal-2019-event} uses Multi-Order Graph Attention Network (GAT) to aggregate multi-order syntactic relations in the sentences based on Stanford CoreNLP parsed POS and syntactic dependency.
    \item \textbf{SS-VQ-VAE} \cite{huang-etal-2020-semi} filters candidate trigger words using an OntoNotes-based Word Sense Disambiguation (WSD) tool and uses BERT for ED.
    \item \textbf{DYGIE++} \cite{wadden-etal-2019-entity} is a multi-task information extraction model which uses BERT and graph-based span population. Gold annotations for ACE 2005 event, entity and relation are all used in training.
\end{itemize}

\noindent Additionally, the highest reported scores for ED on the ACE 2005 data in the literature to date are obtained by:
\begin{itemize}
    \item \textbf{PLMEE} \cite{yang-etal-2019-exploring-pre} is a BERT-based model that is finetuned for ED and argument extraction in a pipelined manner.
\end{itemize}

\noindent We found that PLMEE is trained and evaluated with only eventful sentences from the ACE 2005 dataset. For a fair comparison, we implement SPEED2, which uses the same training and evaluation data as PLMEE.

\subsection{Implementation Details}

We implement the proposed model in Pytorch and use BERT$_{large-uncased}$ with whole word masking. Maximum sequence length is set as 256. We use Adam \cite{Adam} optimizer with the learning rate tuned around 3e-5. The batch size is set between 4-8 to be fit for single-GPU training. We implement early stopping (patience = 5) and limit the training to 50 epochs. We apply dropout of 0.9. For the Label Semantic Learner, we set Transformer encoder/decoder layer N=3, attention heads H=4, and the temperature for Gumbel Sampling $\tau$=0.1.

\begin{table}[t]
\small
\centering
\caption{\label{performance}Performance results for ACE 2005 Event Trigger Classification.}
\resizebox{\textwidth}{!}{%
\begin{tabular}{p{0.35\textwidth}p{0.45\textwidth}ccc}
\hlineB{2}
\textbf{Model} & \textbf{Core mechanism} & \textbf{P} & \textbf{R} & \textbf{F1}\\
\hline
DMCNN \cite{chen-etal-2015-event} & CNN & 75.6 & 63.6 & 69.1 \\
DMBERT \cite{wang-etal-2019-adversarial-training} & Transformer & 77.6&71.8&74.6\\
BERT\_QA \cite{du-cardie-2020-event} & Transformer & 71.1 & 73.7 & 72.4 \\
\hline
JRNN$\ddagger$ \cite{nguyen-etal-2016-joint} & features+RNN & 66.0 & 73.0 & 69.3\\
JMEE$\ddagger$ \cite{liu-etal-2018-jointly} & features+RNN+GCN & 76.3&71.3&73.7\\
MOGANED$\ddagger$ \cite{yan-etal-2019-event} &features+GAT&\textbf{79.5}&72.3&75.7\\
SS-VQ-VAE$\ddagger$ \cite{huang-etal-2020-semi}&WSD+Transformer &75.7&77.8&76.7\\
\hline
DYGIE++$*$ \cite{wadden-etal-2019-entity} & Transformer+Multi-task data & - & - & 73.6\\
DMBERT+Boot$*$ \cite{wang-etal-2019-adversarial-training} & Transformer+Augmented data & 77.9&72.5&75.1\\
\hline
SPEED (ours) & Transformer &76.8&\textbf{77.4}&\textbf{77.1}\\
\hlineB{2}
PLMEE$\dagger$ \cite{yang-etal-2019-exploring-pre} &Transformer&81.0&80.4&80.7\\
SPEED2$\dagger$ (ours) &Transformer& 79.8 & \textbf{86.0} & \textbf{81.4}\\
\hlineB{2}
\multicolumn{5}{l}{\footnotesize Note: The baseline models are grouped by core mechanism: $\ddagger$ indicates the models}\\
\multicolumn{5}{l}{ that use linguistic tools. $*$ indicates those using external resources. $\dagger$ indicates the}\\
\multicolumn{5}{l}{models that are trained and evaluated only on eventful data.}
\end{tabular}

}
\end{table}

\subsection{Experimental Results}

Table~\ref{performance} shows the performance results of our proposed SPEED model based on the ACE 2005 benchmark as compared to the state-of-the-art models. The models are grouped together roughly by their approaches. Without using linguistic tools or external resources, our proposed SPEED model achieves 77.1\% in F1, outperforming the baseline models by 0.4\%-8.0\% in F1. Among all the models, SPEED achieves the highest recall with good precision. Although MOGANED achieves a particularly high precision (79.5\%), its recall is lower than our SPEED model by 5.1\%. One possible reason is that since it utilizes golden entities and syntactic features based on linguistic tools, the model's inductive bias enables it to perform better on the more regular instances. In contrast, SPEED is not based on syntactic prior but semantic prior of the event types. It can cover irregular instances though with less precision.

When trained and evaluated on only eventful data, our SPEED2 outperforms PLMEE in terms of recall (+5.6\%) and F1 (+0.7\%), giving a more balanced performance. The results show that incorporating label information is an effective approach for event detection.

\subsection{Ablation Studies}

\begin{table}[t]
\centering
\caption{\label{table:ablation-2}Ablation study on SPEED.}
\resizebox{.5\textwidth}{!}{%
\begin{tabular}{lccc}
\hline \textbf{Model} & \textbf{F1} & \textbf{$\Delta$ F1} \\
\hline
(1) SPEED model & 77.1 & - \\
(2) TC (large) w/o LSL  &75.2& -1.9\% \\
(3) TC (large) w/o labels as input &73.4& -3.7\% \\
\hline
(4) TC (base) &72.8& -4.3\% \\
(5) TC (base) w/o labels as input  &71.0& -1.8\% \\
\hline
\end{tabular}
}
\end{table}

We conduct ablation experiments to show the effectiveness of the individual components of our model. Table~\ref{table:ablation-2} reports the results in F1: (1) The original SPEED model, whose Trigger Classifier (TC) is based on BERT-large. (2) We remove the Label Semantic Learner (LSL), i.e., the label word representation $L^\prime$ is the same as the original label words $L$. (3) We do not use labels as input, i.e., the LSL is removed and the label word representation $L^\prime$ is not included as part of the input in the TC. (4) We replace BERT-large by BERT-base in the TC. (5) On top of (4), we do not use labels as input for the model, similar to (3).

The results show that all the key components in our proposed SPEED model are necessary and effective for ED. Firstly, we observe that removing LSL significantly reduces performance by 1.9\%. Secondly, replacing TC (large) with TC (base) leads to a drop in performance by 4.3\%. This is possibly because the BERT-large provides better contextual word representation and more space for interaction between a sentence and type semantic pivot words than its base counterpart. Finally, regardless of the BERT version used in the Trigger Classifier, performance degradation is significant if the event type labels are not used as input to provide the semantic prior information.

\subsection{Analysis and Discussion}
\label{sec:analysis}

\begin{figure}[t!]
\centering
\includegraphics[scale=.35]{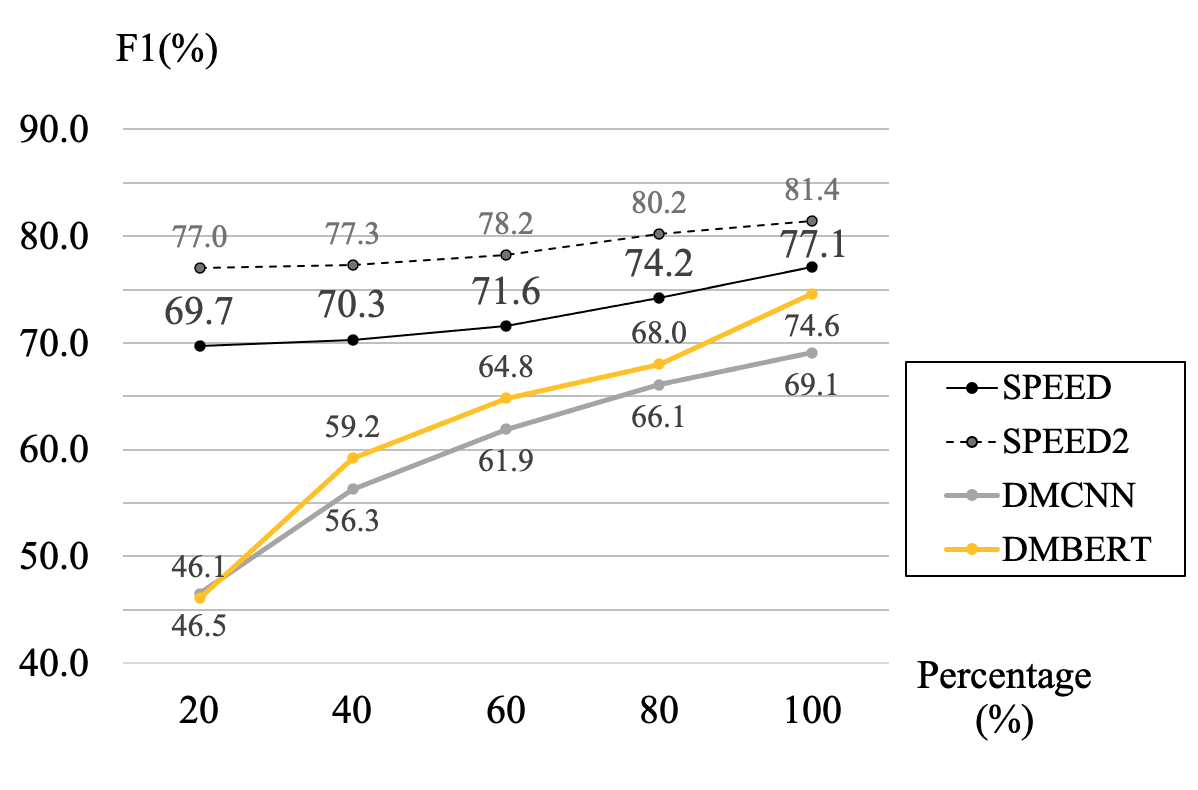}
\caption{Performance on scarce training data.} \label{fig:less}
\end{figure}

\paragraph{Analysis on Scarce Training Data Scenario Performance.} To show the data efficiency of our proposed SPEED model, we also evaluate it on scarce training data in comparison with the baseline models DMCNN and DMBERT. More specifically, we evaluate the models after training them with 20\%, 40\%, 60\% and 80\% of the training data. As shown in Figure~\ref{fig:less}, our model performs significantly better than the baselines under the settings. With less training data, the performance of DMCNN and DMBERT drops significantly by 3.0\%-28.5\% in F1, while the performance of our proposed SPEED model only drops by 2.9\%-7.4\%. With an extremely limited amount (20\%) of data, SPEED can still achieve a reasonable F1 performance of 69.7\%. In the same setting, DMCNN and DMBERT can only achieve around 46\% in F1. Similar to SPEED, SPEED2 which is evaluated with only eventful sentences shows reasonable performance degradation with significantly reduced amount of training data. This shows the effectiveness of the proposed model in learning from scarce data for ED.

\begin{table}[t]
\centering
\caption{Performance comparison on single (1/1) and multiple (1/N) event sentences.}\label{table:inter-event}
\resizebox{.5\textwidth}{!}{%
\begin{tabular}{>{\arraybackslash}p{3cm}>{\centering\arraybackslash}p{1cm}>{\centering\arraybackslash}p{1cm}>{\centering\arraybackslash}p{1cm}}
\hline \textbf{Model} &\textbf{1/1} & \textbf{1/N}& \textbf{All} \\
\hline
DMCNN & 74.3&50.9&69.1\\
JRNN & 75.6&64.8&69.3 \\
JMEE &75.2&72.7&73.7\\
\hline
SPEED (ours) &\textbf{77.5}&\textbf{76.8}&\textbf{77.1}\\
\hline
\end{tabular}
}
\end{table}

\paragraph{Analysis on Single/Multiple Event Sentence Performance.} Among all the baselines, JMEE focuses on addressing multiple event sentences, i.e., sentences with each containing more than one event trigger. In Table~\ref{table:inter-event}, we report our F1 performance on single event sentences and multiple event sentences, in comparison with JMEE and the strong baselines it used for this scenario (i.e., DMCNN and JRNN). Without using linguistic features including POS tag and dependency, our SPEED model achieves high F1 (76.8\%) performance on multiple event sentences, outperforming the baselines by 4.1\%-25.9\%. It shows that our proposed architecture can effectively model cross-event interaction, benefiting ED on multiple event sentences.

\section{Conclusion}
\label{sec:conclusion}

In this paper, we propose a novel semantic pivoted Event Detection model that utilizes the pre-defined set of event type labels for event detection. It features event type semantics learning via a Transformer-based mechanism. The experimental results show that our model outperforms the state-of-the-art event detection methods. In addition, the proposed model demonstrates several other advantages, such as working well for the scenarios of scarce training data and multiple event sentences.

%
%
%
\bibliographystyle{splncs04}
%
\bibliography{example}

\end{document}